\documentclass{article}
\usepackage{spconf,url,bm,amssymb,amsmath,graphicx}
\usepackage{algorithm,algorithmicx,algpseudocode}
\usepackage{balance}

\title{END-TO-END SPEECH RECOGNITION WITH WORD-BASED RNN LANGUAGE MODELS}
\name{Takaaki Hori$^1$, Jaejin Cho$^2$, Shinji Watanabe$^2$}
\address{$^1$Mitsubishi Electric Research Laboratories (MERL)\\
$^2$Center for Language and Speech Processing, Johns Hopkins University\\
thori@merl.com, \{jcho52, shinjiw\}@jhu.edu}
\begin{document}
\ninept
\maketitle
\begin{abstract}
This paper investigates the impact of word-based RNN language models (RNN-LMs) on the performance of end-to-end automatic speech recognition (ASR). In our prior work, we have proposed a multi-level LM, in which character-based and word-based RNN-LMs are combined in hybrid CTC/attention-based ASR.
Although this multi-level approach achieves significant error reduction in the Wall Street Journal (WSJ) task, two different LMs need to be trained and used for decoding, which increase the computational cost and memory usage.
In this paper, we further propose a novel word-based RNN-LM, which allows us to decode with only the word-based LM, where it provides look-ahead word probabilities to predict next characters instead of the character-based LM, leading competitive accuracy with less computation compared to the multi-level LM.
We demonstrate the efficacy of the word-based RNN-LMs using a larger corpus, LibriSpeech, in addition to WSJ we used in the prior work.
Furthermore, we show that the proposed model achieves 5.1 \%WER for WSJ Eval'92 test set when the vocabulary size is increased, which is the best WER reported for end-to-end ASR systems on this benchmark.
\end{abstract}
\section{Introduction}
\label{sec:intro}
Automatic speech recognition (ASR) is currently a mature set of widely-deployed technologies that enable successful user interface applications such as voice search \cite{sainath-icassp-2015}.
However, current systems lean heavily on the scaffolding of complicated legacy architectures that grew up around traditional techniques, including hidden Markov models (HMMs), Gaussian mixture models (GMMs), hybrid HMM/deep neural network (DNN) systems, and sequence discriminative training methods \cite{Povey_ASRU2011}.
These systems also require hand-made pronunciation dictionaries based on linguistic assumptions, extra training steps to derive context-dependent phonetic models, and text preprocessing such as tokenization for languages without explicit word boundaries.
Consequently, it is quite difficult for non-experts to develop ASR systems for new applications, especially for new languages.  

End-to-end ASR has the goal of simplifying the above module-based architecture into a single-network architecture within a deep learning framework, in order to address these issues. End-to-end ASR methods typically rely only on paired acoustic and language data without linguistic knowledge, and train the model with a single algorithm. Therefore, the approach makes it feasible to build ASR systems without expert knowledge.
There are several types of end-to-end architecture for ASR such as connectionist temporal classification (CTC) \cite{graves2014towards}, recurrent neural network (RNN) transducer \cite{graves2013speech}, attention-based encoder decoder \cite{chorowski2015attention}, and their hybrid models \cite{kim2016joint_icassp2017, hori-ACL-2017}.

Recently, the use of external language models has shown significant improvement of accuracy in neural machine translation \cite{gulcehre2015using} and end-to-end ASR \cite{hori-interspeech-2017, kannan2017analysis}.
This approach is called {\it shallow fusion}, where the decoder network is combined with an external language model in log probability domain for decoding.
In our previous work \cite{hori-interspeech-2017}, we have shown the impact of recurrent neural network language models (RNN-LMs) in Japanese and Mandarin Chinese tasks, reaching a comparable or higher accuracy to those of state-of-the-art DNN/HMM systems. 
Since the Japanese and Chinese systems were designed to output character sequences, the RNN-LM was also designed as a character-based LM, and effectively combined with the decoder network to jointly predict the next character.

A character-based architecture achieves high-accuracy ASR for languages with a large set of characters such as Japanese and Chinese. It also enables open vocabulary ASR, in contrast to word-based architectures, which suffer from the out-of-vocabulary (OOV) problem.
However, the character-based LMs generally under-perform relative to word LMs for languages with a phonogram alphabet using fewer distinct characters, such as English, because of the difficulty of modeling linguistic constraints across long sequences of characters. Actually, English sentences are much longer than Japanese and Chinese sentences in the length of character sequence.
To overcome this problem, we have further extended end-to-end ASR decoding with LMs at both the character and word levels \cite{hori2017multi}. During the beam search decoding, Hypotheses are first scored with the character-based LM until a word boundary is encountered. Known words are then re-scored using the word-based LM, while the character-based LM provides for out-of-vocabulary scores. This approach exploits the benefits of both character and word level architectures, and enables high-accuracy open-vocabulary end-to-end ASR.

More specifically, the character-based LM yields the following benefits in the decoding process with the word-based LM:
\begin{enumerate}
    \item Character-based LM can help correct hypotheses survive until they are rescored at word boundaries during the beam search. Before the hypothesis reaches the boundary, the identity of the last word is unknown and its word probability cannot be applied. Hence, good character-level prediction is important to avoid pruning errors for hypotheses within a word. 
    \item Character-based LM can predict character sequences even for OOV words not included in the vocabulary of the word-based LM. Since the word-based LM basically cannot predict unseen character sequences, good character-level prediction is important for open-vocabulary ASR.
\end{enumerate}

However, the multi-level LM approach has a problem that it requires two different RNN-LMs.
To build the two LMs, we need to take additional time and effort, almost twice of them, for training the models.
Moreover, the two LMs also increase the computational cost and memory usage for decoding.
Inherently, RNN-LMs need a lot of computation for training and decoding compared with conventional $N$-gram LMs.
In addition, text corpora for training LMs are usually much larger than paired acoustic and text data for training end-to-end ASR models. Considering this situation, solving the above problem is crucial for better end-to-end ASR.
    
In this paper, we propose a novel strategy for language modeling and decoding in end-to-end ASR to solve the problem.
The proposed method allows us to decode with only a word-based RNN-LM in addition to the encoder decoder, leading a competitive accuracy and less computation in the decoding process compared to the multi-level LM approach.
This method employs look-ahead word probabilities to predict next characters instead of the character-based LM.
Although our approach is similar to old fashioned lexical-tree search algorithms including language model look-ahead \cite{ortmanns1996language, alleva1996improvements}, it provides an efficient way of dynamically computing the look-ahead probabilities for end-to-end ASR with a word-based RNN-LM, which does not exist in the prior work.
We demonstrate the efficacy of the proposed LMs on standard Wall Street Journal (WSJ) and LibriSpeech tasks. 

\section{End-to-end ASR architecture}
\label{sec:joint-ctc-attention}
This section explains the hybrid CTC/attention network \cite{kim2016joint_icassp2017,hori-ACL-2017} we used for evaluating the proposed language modeling and decoding approach. 
But the proposed LMs can also be applied to standard attention-based encoder decoders for ASR.

\subsection{Network architecture}
\label{sec:network_architecture}
\begin{figure}[tb]
	\includegraphics[width=8.3cm]{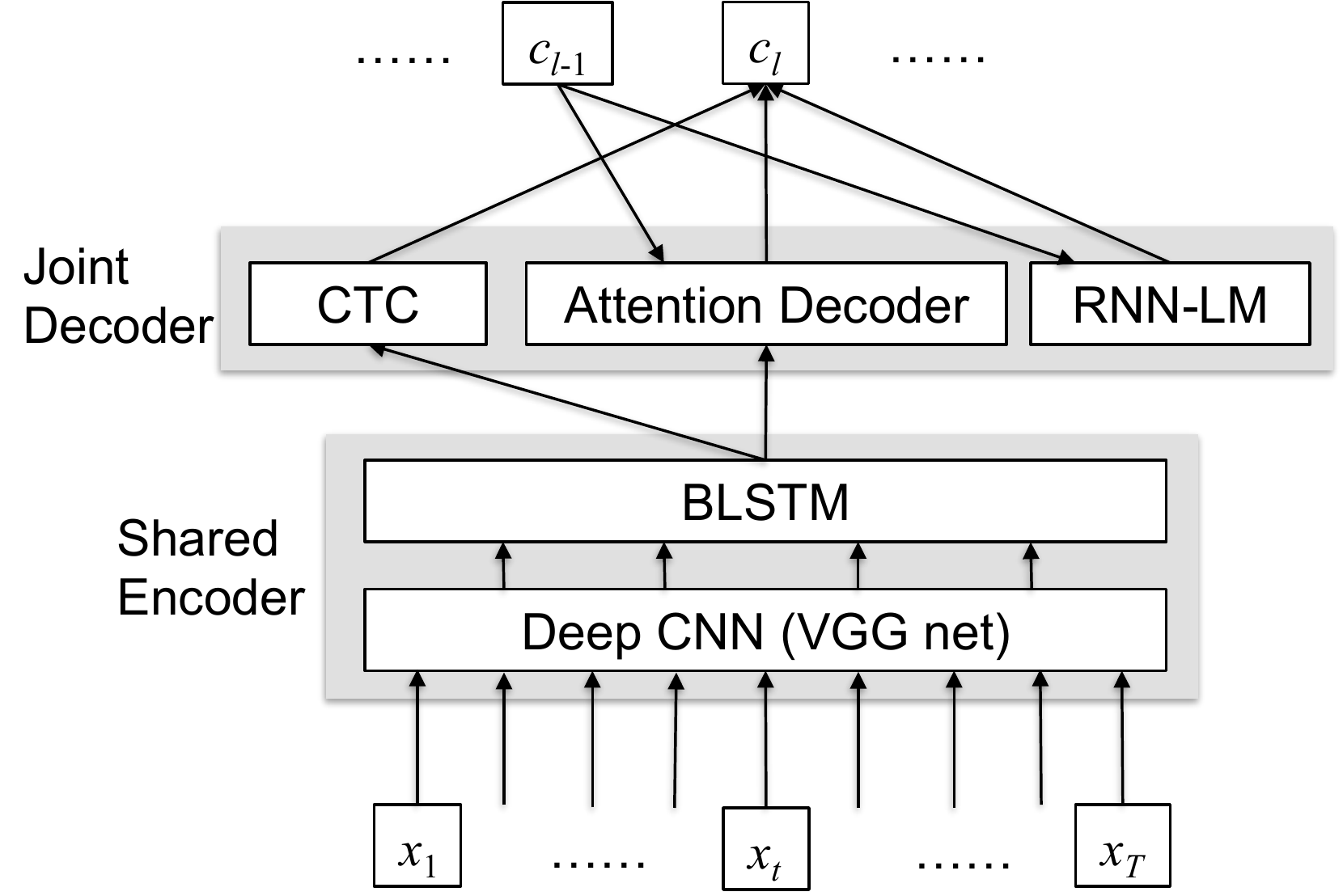}
	\caption{Hybrid attention/CTC network with LM extension: the shared encoder contains a VGG net followed by BLSTM layers and trained by both CTC and attention model objectives simultaneously. The joint decoder predicts an output label sequence by the CTC, attention decoder and RNN-LM.}
	\label{fig:ctc-attention}
\end{figure}

Figure \ref{fig:ctc-attention} shows the latest architecture of the CTC/attention network \cite{hori-interspeech-2017}. 
The encoder has deep convolutional neural network (CNN) layers with the VGG net architecture \cite{simonyan2014very}, which are followed by stacked bidirectional long short-term memory (BLSTM) layers.
The decoder network has a CTC network, an attention decoder network, and an RNN-LM, which jointly predict the next label.
Given input sequence $X=x_1,\dots,x_T$, the encoder network accepts $X$ and outputs hidden vector sequence $\mathbf{H}=\mathbf{h}_1,\dots,\mathbf{h}_{T'}$, where $T'=T/4$ by using two max-pooling steps in the deep CNN.
The decoder network iteratively predicts a single label $c_l$ based on the hidden vectors $\mathbf{H}$ and the label context $c_1,\dots,c_{l-1}$, and generates $L$-length label sequence $C = \{ c_l \in \mathcal{U} | l = 1, \cdots, L\}$, where $\mathcal{U}$ is a set of labels.
In this work, we assume $\mathcal{U}$ is a set of distinct characters or alphabet of the target language.

The hybrid attention/CTC network utilizes both benefits of CTC and attention during training and decoding by sharing the same CNN/ BLSTM encoder with CTC and attention decoder networks and training them jointly.
Unlike the solitary attention model, the forward-backward algorithm of CTC can enforce monotonic alignment between speech and label sequences during training.
 That is, rather than solely depending on the data-driven attention mechanism to estimate the desired alignments in long sequences, the forward-backward algorithm in CTC helps to speed up the process of estimating the desired alignment.
The objective to be maximized is a logarithmic linear combination of the CTC and attention-based posterior probabilities $p_\text{ctc} (C|X)$ and $p_\text{att} (C|X)$: 
\begin{align}
\label{e12}
\mathcal{L}_\text{MTL} &= \lambda \log p_\text{ctc} (C|X) + (1-\lambda) \log p_\text{att} (C|X),
\end{align}
with a tunable parameter $\lambda: 0 \leq \lambda \leq 1$.

\subsection{Decoding with external language models}
\label{sec:decoding}
The inference step of CTC/attention-based speech recognition is performed by output-label synchronous decoding with a beam search.
Although the decoding algorithm is basically the same as the method for standard attention-based encoder decoders, it also considers the CTC and LM probabilities to find a better hypothesis.
The decoder finds the most probable character sequence $\hat{C}$ given speech input $X$, according to
\begin{align}
\hat{C}=\arg \max_{C\in \mathcal{U}^*} & \left\{\lambda \log p_{\text{ctc}}(C|X) + (1-\lambda)\log p_{\text{att}}(C|X)\right. \nonumber \\
& \left.  + \gamma \log p_\text{lm}(C)\right\},
\label{eq:lm_decoding}
\end{align}
where LM probability $p_\text{lm}(C)$ is added with scaling factor $\gamma$ to the CTC/attention probability in the log probability domain.

In the beam search process, the decoder computes a score of each partial hypothesis, which is defined as the log probability of the hypothesized character sequence. 
The joint score $\alpha(g)$ of each partial hypothesis $h$ is computed by
\begin{align}
\alpha(h)=\lambda \alpha_\text{ctc}(h) + (1-\lambda)\alpha_\text{att}(h) + \gamma \alpha_\text{lm}(h),
\end{align}
where $\alpha_\text{ctc}(h)$, $\alpha_\text{att}(h)$, and $\alpha_\text{lm}(h)$ are CTC, attention, and LM scores, respectively.

With the attention model, score $\alpha_\text{att}(h)$ can be obtained recursively as
\begin{align}
\alpha_\text{att}(h)=\alpha_\text{att}(g) + \log p_\text{att}(c|g,X),
\label{eq:score_recursion}
\end{align}
where $g$ is an existing partial hypothesis, and $c$ is a character label appended to $g$ to generate $h$, i.e., $h=g\cdot c$. 
The score for $h$ is obtained as the addition of the original score $\alpha_\text{att}(g)$ and the conditional log probability given by the attention decoder.
LM score $\alpha_\text{lm}(h)$ is also obtained similarly to the attention model as
\begin{align}
\alpha_\text{lm}(h)=\alpha_\text{lm}(g) + \log p_\text{lm}(c|g).
\end{align}

On the other hand, CTC score $\alpha_\text{ctc}(h)$ is obtained differently from the other scores, where we compute the CTC prefix probability \cite{graves2008thesis} defined as the cumulative probability of all label sequences that have $h$ as their prefix: 
\begin{align}
p(h,\dots|X)=\sum_{\nu \in (\mathcal{U}\cup \{\text{\tt <eos>}\})^+} P(h \cdot \nu|X),
\end{align}
and use it as the CTC score:
\begin{align}
\alpha_\text{ctc}(h) \triangleq \log p(h,\dots|X),
\end{align}
where $\nu$ represents all possible label sequences except the empty string, and {\tt <eos>} indicates the end of sentence. 

During the beam search, the number of partial hypotheses for each length is limited to a predefined number, called a {\it beam width}, to exclude hypotheses with relatively low scores, which dramatically improves the search efficiency.

\section{Incorporating Word-based RNN-LMs}
In this section, we explain the basic approach to incorporate word-based LMs into a character-based end-to-end ASR, and present two word-based RNN-LMs, one is a multi-level LM we have already proposed and the other is a look-ahead word LM we propose in this paper.

\subsection{Basic approach}
In most end-to-end ASR systems, a finite lexicon and an $N$-gram language model are compiled into a Weighted Finite-State Transducer (WFST), and used for decoding \cite{miao2015eesen,bahdanau2016end}.
The WFST framework efficiently handles frame-synchronous or label-synchronous decoding with the optimized search network and reduces the word error rate \cite{mohri2002weighted,hori2013speech}.
However, this approach is not suitable for RNN-LMs because an RNN-LM cannot be represented as a static state network.

In this paper, we extend the character-based decoding to enable open-vocabulary end-to-end ASR with a word-based RNN-LM.
We consider that the character-based systems can predict space characters between words as well as letters within the word. Note that the space character has an actual character code, which is different from the CTC's blank symbol.
With the space characters, it is possible to deterministically map any character sequence to a word sequence, e.g., character sequence
\begin{center}
\verb+a <space> c a t <space> e a t s+
\end{center}
is mapped to a unique word sequence
\begin{center}
\verb+a cat eats+
\end{center}
where {\tt <space>} formally represents the space character.
Accordingly, only when the decoder hypothesizes a space character, it computes the probability of the last word using the word-level RNN-LM and simply accumulates it to the hypothesis score.  No special treatment is necessary for different types of homonyms: words with the same spelling but different pronunciation are handled in a context-dependent way by the word language model, whereas words with the same pronunciation but different spellings are automatically handled as different word hypotheses in the beam search.  Similarly, ambiguous word segmentations are automatically handled as different decoding hypotheses. 

\subsection{Multi-level RNN-LM}
The multi-level RNN-LM contains character-level and word-level RNN-LMs, but it can be implemented as a function that performs character-level prediction.
Let $\cal V$ be the vocabulary of the word-level RNN-LM and be including an abstract symbol of OOV word such as {\tt <UNK>}.
We compute the conditional character probabilities with
\begin{align}
p_\text{lm}(c|g) = & \left\{
\begin{array}{ll}
\frac{p_\text{wlm}(w_g|\psi_g)}{p_\text{clm}(w_g|\psi_g)} & \text{if}~c\in S, w_g\in {\cal V}\\
p_\text{wlm}(\text{\tt <UNK>}|\psi_g) \tilde{\beta} & \text{if}~c \in S, w_g\not\in {\cal V} \\
p_\text{clm}(c|g)                            & \text{otherwise} \\
\end{array}
\right.
\label{eq:lmprob}
\end{align}
where $S$ denotes a set of labels that indicate the end of word, i.e.,  $S=\{\text{\tt <space>},\text{\tt <eos>}\}$, $w_g$ is the last word of the character sequence $g$, and $\psi_g$ is the word-level history, which is the word sequence corresponding to $g$ excluding $w_g$.
For the above example, $g$, $w_g$, and $\psi_g$ are set as
\begin{align}
g&=\text{\tt a}, \text{\tt <space>}, \text{\tt c},\text{\tt a}, \text{\tt t}, \text{\tt <space>}, \text{\tt e}, \text{\tt a}, \text{\tt t}, \text{\tt s} \nonumber \\
w_g&=\text{\tt eats} \nonumber \\
\psi_g&=\text{\tt a}, \text{\tt cat} \nonumber.
\end{align}
$\tilde{\beta}$ is a scaling factor used to adjust the probabilities for OOV words.

The first condition on the right-hand side of Eq. \eqref{eq:lmprob} is applied when the $g$ has reached the end of a word.
In this case, the word-level probability $p_\text{wlm}(w_g|\psi_g)$ is computed using the word-level RNN-LM.
The denominator $p_\text{clm}(w_g|\psi_g)$ is the probability of $w_g$ obtained by the character-level RNN-LM and used to cancel the character-level LM probabilities accumulated for $w_g$. 
The probability can be computed as
\begin{align}
p_\text{clm}(w_g|\psi_g)=\prod_{i=1}^{|w_g|} p_\text{clm}(w_{g,i}|\psi_g w_{g,1} \cdots w_{g,i-1}),
\end{align}
where $|w_g|$ is the length of word $w_g$ in characters and $w_{g,i}$ indicates the $i$-th character of $w_g$.  
The second term, $ p_\text{wlm}(\text{\tt <UNK>}|\psi_g)$ acts as a weight on the character-level LM and ensures that the combined language model is normalized over character sequences both at word boundaries and in-between.  
If $w_g$ is an OOV word as in the second condition, we assume that a word-level probability for the OOV word can be computed with the word and character-level RNN-LMs as
\begin{align}
p_\text{oov}(w_g|\psi_g) & = p_\text{wlm}(\text{\tt <UNK>}|\psi_g) p_\text{clm}(w_g|\text{\tt <UNK>},\psi_g). 
\end{align}
Since the character-level probability satisfies
\begin{align}
p_\text{clm}(w_g|\text{\tt <UNK>},\psi_g) \propto p_\text{clm}(w_g|\psi_g)
\end{align}
and
\begin{align}
p_\text{clm}(w_g|\text{\tt <UNK>},\psi_g) = \beta(\psi_g)~p_\text{clm}(w_g|\psi_g), 
\end{align}
we approximate $\beta(\psi_g) \approx \tilde{\beta}$ and use $\tilde{\beta}$ as a tunable parameter.
In the second condition of Eq.~\eqref{eq:lmprob}, character-based probability $p_\text{clm}(w_g|\psi_g)$ is eliminated since it is already accumulated for the hypothesis.

The third case gives the character-level LM probabilities to the hypotheses within a word. Although the character-level LM probabilities are canceled at the end of every known word hypothesis and so are only used to score OOV words, they serve another important role in keeping the correct word hypotheses active in the beam search until the end of the word where the word-level LM probability is applied.   

Finally, the log probability of sentence-end label {\tt <eos>} is added to the log probability of each complete hypothesis $g'$ as
\begin{align}
\alpha(g')=\alpha(g) + \gamma \log p_\text{wlm}(\text{\tt <eos>}|\psi_g w_g)
\end{align}
in the beam search process.

\subsection{Look-ahead word-based RNN-LM}
The look-ahead word-based RNN-LM enables us to decode with only a word-based RNN-LM in addition to the encoder decoder.
This model predicts next characters using a look-ahead mechanism over the word probabilities given by the word-based LM, while the multi-level LM uses a character-level LM until the identity of the word is determined. 

\begin{figure}[tb]
	\includegraphics[width=8cm]{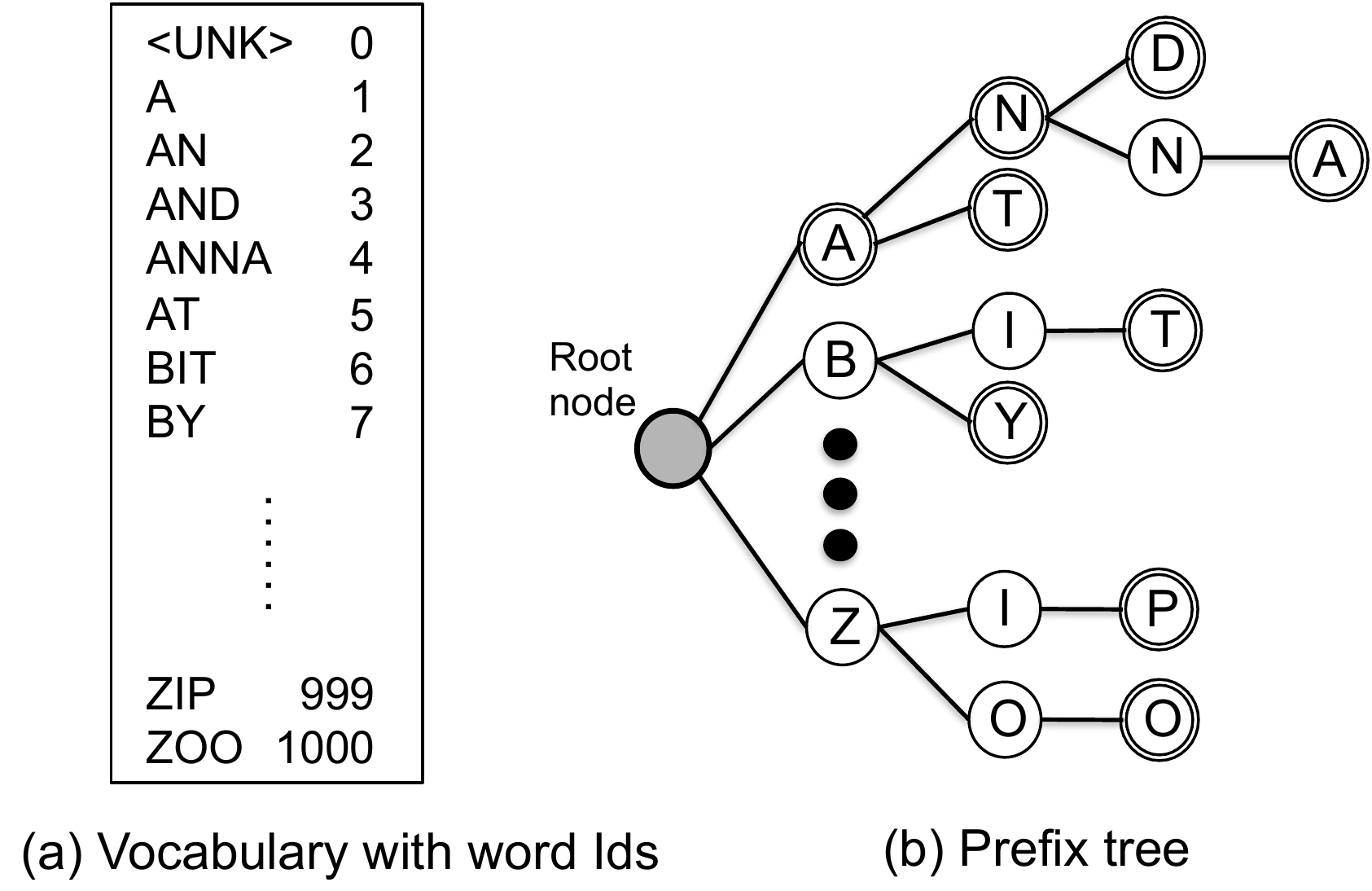}
	\caption{Prefix tree representation of a vocabulary. (a) vocabulary including word strings and id numbers. (b) prefix tree of the vocabulary, where the shaded circle indicates the root node, each white circle is a node representing a character and each path from the root node represents a character sequence of each word, where each double-circle node corresponds to a word end.}
	\label{fig:prefix-tree}
\end{figure}

To compute look-ahead probabilities efficiently, we use a prefix tree representation as shown in Fig.~\ref{fig:prefix-tree}.
This example shows a vocabulary and its prefix tree representation.
During decoding, each hypothesis holds a link to a node, which indicates where the hypothesis is arriving in the tree.
Suppose a set of anticipated words at each node has already been obtained in advance.
A look-ahead probability at node $n$ can be computed as the sum of the word probabilities of all the anticipated words as
\begin{align}
    p_\text{la}(n|\psi) = \sum_{w \in \text{wset}(n)} p_\text{wlm}(w|\psi),
    \label{eq:lookahead}
\end{align}
where $\text{wset}(n)$ denotes the set of anticipated words at node $n$, and $p_\text{wlm}(w|\psi)$ is the original word probability given by the underlying word-based RNN-LM for word-level context $\psi$.

The character-based LM probability with the look-ahead mechanism is computed as
\begin{align}
p_\text{lm}(c|g) = & \left\{
\begin{array}{ll}
    p_\text{wlm}(w_g|\psi_g) / p_\text{la}(n_g|\psi_g) & \text{if}~ n_g \in F, c\in S \\
    p_\text{la}(n_{g\cdot c}|\psi_g) / p_\text{la}(n_g|\psi_g) & \text{if}~ n_g \neq \text{null}, c\in \xi(n_g) \\
    p_\text{wlm}(\text{\tt <UNK>}|\psi_g) \eta & \text{if}~ n_g \neq \text{null}, c \not \in \xi(n_g) \\
    1                    & \text{otherwise} \\
\end{array}
\right.
\label{eq:lookahead_charprob}
\end{align}
where $F$ denotes a set of word end nodes, $n_g$ is the node that $g$ has arrived, $n_{g \cdot c}$ is a succeeding node of $n_g$ determined by $c$, $\xi(n_g)$ is a set of succeeding nodes from $n_g$, and $\eta$ is a scaling factor for OOV word probabilities, which is a tunable parameter.

The first case of Eq.~\eqref{eq:lookahead_charprob} gives the word probability at a word-end node, where $p_\text{wlm}(w_g|\psi_g)$ needs to be normalized by $p_\text{la}(n_g|\psi_g)$ to cancel the already accumulated look-ahead probabilities.
The second case computes the look-ahead probability when making a transition from node $n_g$ to $n_{g \cdot c}$. The third case gives the OOV word probability, where character $c$ is not accepted, which means the hypothesis is going to an OOV word.
The last one handles the case that $n_g$ is null, which means that the hypothesis is already out of the tree, and it returns 1 since the OOV probability was already applied in the third case.
In the above procedure, we assume that whenever the hypothesis is extended by {\tt <space>} label, the new hypothesis points the root node of the tree.

Although this approach is similar to conventional ASR systems based on a prefix tree search including a LM look-ahead mechanism, the look-ahead probabilities needs to be computed on the fly using the word-based RNN-LM unlike conventional approaches with unigram or bigram LM or weight pushing over a static WFST.

To compute the sum of probabilities in Eq.~\eqref{eq:lookahead}, we assume that the Id numbers are assigned in alphabetical order in the vocabulary. In this case, the Id numbers should be consecutive in each set as a property of prefix tree.
Accordingly, we can compute the sum using the cumulative sums over the word probability distribution by 
\begin{align}
    p_\text{la}(n|\psi) = s_{\psi}[\text{max\_id}(n)]-s_{\psi}[\text{min\_id}(n)-1],
    \label{eq:cumsum_diff}
\end{align}
where $s_{\psi}[\cdot]$ denotes an array of the cumulative sums given context $\psi$, which is obtained as 
\begin{align}
    s_{\psi}[i] = \sum_{k=0}^{i} p_\text{wlm}(w^{(k)}|\psi) ~~~\text{for}~i=0,\dots,|{\cal V}|,
\end{align}
and $\text{max\_id}(n)$ and $\text{min\_id}(n)$ are the maximum and minimum Id numbers in the set of anticipated words at node $n$. $w^{(k)}$ denotes the $k$-th word in the vocabulary.
Once the cumulative sums are computed right after the softmax operation, we can quickly compute the look-ahead probabilities.

\section{Related work}
There are some prior work, that incorporates word units into end-to-end ASR. One major approach is acoustic-to-word CTC \cite{soltau2016neural, audhkhasi2017direct, li2017acoustic}, where the input acoustic feature sequence is directly mapped to the word sequence using CTC.
However, this approach essentially requires a large amount of paired acoustic and text data to learn acoustic mapping to a large number of words. For example, \cite{soltau2016neural} used 125,000 hours of transcribed audio data to train the word CTC model.

Our approach, in contrast, is specially designed for end-to-end ASR using a character or subword-based encoder decoder and an external RNN language model trained with a large text corpus.
Thus, this architecture is more suitable for low-resource languages, where the amount of parallel data is limited but large text data are available.

Subword units \cite{wu2016google,rao2017exploring} are also available as an intermediate representation of character and word, where the unit set is automatically obtained by byte-pair encoding \cite{gage1994new} or some chunking techniques.
However, this approach needs to select an appropriate number of subword units using training data. Increasing the number of units will lead more acoustically expressive units but make it more difficult to train the encoder decoder using a limited amount of data.
In addition, it assumes to use the subword units for both the encoder decoder and the language model, but the appropriate units can be different for the encoder decoder and the language model.
Our approach basically employs characters and words, but it is also possible to combine a character-based encoder decoder with a subword-based LM or a subword-based encoder decoder with a word-based LM.

\section{Experiments}
We evaluate the proposed language models with the Wall Street Journal (WSJ) and LibriSpeech corpora.
WSJ is a well-known English clean speech database \cite{wsj1, garofalo2007csr} including approximately 80 hours data while
LibriSpeech is a larger data set of read speech from audiobooks, which contains 1000 hours of audios and transcriptions \cite{panayotov2015librispeech}.
\subsection{Evaluation with WSJ}
We used the si284 data set for training, the dev93 data set for validation, and the eval92 data set for evaluation. The data sets are summarized in Table \ref{table:wsj}.
\begin{table}[h]
\caption{WSJ data sets used for evaluation}
\label{table:wsj}
\begin{center}
\begin{tabular}{c|c|c}
    \hline
           & \# utterances & Length (h) \\
    \hline
    Training (WSJ1 si284) & 37,416 & 80 \\
    Validation (dev93) & 503 & 1.1  \\
    Evaluation (eval92) & 333 & 0.7  \\
    \hline
\end{tabular}
\end{center}
\end{table}

As input features, we used 80 mel-scale filterbank coefficients with pitch features and their delta and delta delta features for the CNN/BLSTM encoder  \cite{zhang2016very}.
Our encoder network is boosted by using deep CNN, which is discussed in Section \ref{sec:network_architecture}.
We used a 6-layer CNN architecture based on the initial layers of the VGG-like network \cite{simonyan2014very} followed by eight BLSTM layers in the encoder network.
In the CNN architecture, the initial three input channels are composed of the spectral features, delta, and delta delta features.
Input speech feature images are downsampled to $(1/4 \times 1/4)$ images along with the time-frequency axes through the two max-pooling layers.
The BLSTM layers had 320 cells in each layer and direction, and the linear projection layer with 320 units is followed by each BLSTM layer. 
 
We used the location-based attention mechanism \cite{chorowski2015attention}, where the 10 centered convolution filters of width 100 were used to extract the convolutional features.
The decoder was a one-layer unidirectional LSTM with 300 cells.
We used only 32 distinct labels: 26 English letters, apostrophe, period, dash, space, noise, and sos/eos tokens. 

The AdaDelta algorithm \cite{zeiler2012adadelta} with gradient clipping \cite{pascanu2012difficulty} was used for the optimization.
We also applied a unigram label smoothing technique \cite{chorowski2016towards} to avoid over-confidence predictions.
In the hybrid attention/CTC architecture, we used the $\lambda = 0.1$ for training and the $\lambda=0.2$ for decoding.
The beam width was set to 30 in decoding under all conditions.
The joint CTC-attention ASR was implemented by using the Chainer deep learning toolkit \cite{tokui2015chainer}.

Character and word-based RNN-LMs were trained with the WSJ text corpus, which consisted of 37M words from 1.6M sentences.
The character-based LM had a single LSTM layer with 800 cells and a 32-dimensional softmax layer while the word-based LM had a single LSTM layer with 1000 cells.
We trained word-based RNN-LMs for 20K, 40K and 65K vocabularies,
where the softmax layer had 20K, 40K or 65K-dimensional output in each LM. We used the stochastic gradient descent (SGD) to optimize the RNN-LMs.

The first experiment evaluates the contributions of language models.
Table \ref{tb:wsj_wer} shows word error rate (WER) with different language models. 
\begin{table}[tb]
	\centering
		\caption{Word error rate (WER) with different language models on WSJ.}
		\label{tb:wsj_wer}
		\begin{tabular}{l||c|c}
		     \hline
		     Language models (vocab. size) & dev93 & eval92 \\
		     \hline
		      No LM             & 17.3 & 13.4 \\
		      Chararacter LM    & 12.3 & 7.7  \\
              Word LM (20K)     & 17.1 & 12.6 \\
		      Multi-level LM (20K) & 9.6 & 5.6 \\
		      Look-ahead LM (20K) & 9.5 & 6.1 \\
		      Multi-level LM (40K) & 9.3 & 5.3 \\
		      Look-ahead LM (40K) & 8.6 & 5.3 \\
		      Multi-level LM (65K) & 9.0 & 5.4 \\
		      Look-ahead LM (65K) & {\bf 8.4} & {\bf 5.1} \\
		      \hline
		\end{tabular}
\end{table}
The WERs for no language model (No LM), character LM, word LM (20K) and multi-level LM (20K) were already reported in \cite{hori2017multi}, where the multi-level LM (20K) performed the best in our prior work. 
When we simply applied the word-based LM without any character-level LMs or look-ahead mechanism, the WER reduction was very small due to the pruning errors discussed in Introduction. 

After that, we conducted experiments with the look-ahead LM (20K), where the WER for eval92 test set increased from 5.6 \% to 6.1 \%.
We analyzed the recognition results, and found that the increased errors mainly came from OOV words. This could be because the look-ahead LM did not use a strong character-based LM for predicting OOV words.
To mitigate this problem, we increased the vocabulary size to 40K and 65K. Then, we obtained a large improvement reaching 5.1 \% WER for the look-ahead LM.
The reason why the 65K-look-ahead LM achieved lower WERs than those of multi-level LMs is probably that the look-ahead mechanism provided better character-level LM scores consistent with the word LM probabilities, which were helpful in the beam search process.

\begin{figure}[tb]
	\includegraphics[width=8.5cm]{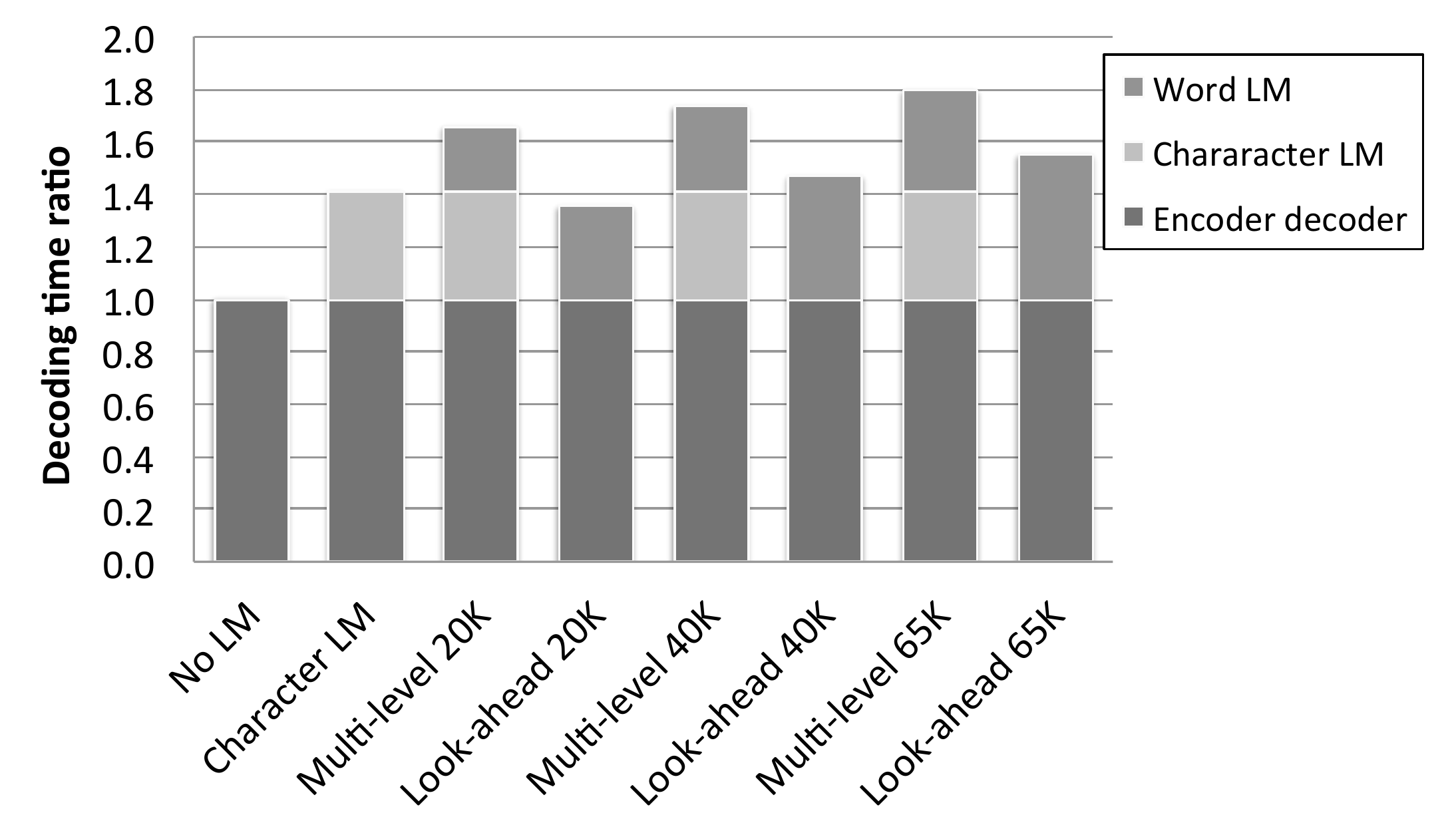}
	\caption{Decoding time ratio in the beam search using different language models. Every elapsed time was divided by that of the No LM case.}
	\label{fig:decoding-time}
\end{figure}

Next, we investigated the decoding time when using different language models.
Figure \ref{fig:decoding-time} shows the decoding time ratio, where each decoding time was normalized by the pure end-to-end decoding time without language models, i.e., the case of No LM, where we only used a single CPU for each decoding process \footnote{Since the beam search-based decoding was implemented in Python, the decoding speed has not been optimized sufficiently.}.
The character LM and 20K-word multi-level LM increased the decoding time by 40\% and 65\%, respectively, while the 20K-word look-ahead LM increased it by 38\%.
Even when we used the 65K-word LMs, the decoding time for the look-ahead LM was still 55\%, which was less than that of the 20K multi-level LM.
Thus, the proposed look-ahead LM has a higher accuracy to multi-level LMs with less decoding time.

Finally, we compare our result with other end-to-end systems reported on the WSJ task. Table \ref{tb:wsj_wer2} summarizes the WER numbers obtained from other articles and this work. Since the systems in the table have different network architectures from each other, it is difficult to compare these numbers directly. However, we confirmed that our system has achieved the best WER in the state-of-the-art systems on the WSJ benchmark.
\begin{table}[tb]
	\centering
		\caption{Comparison with other end-to-end ASR systems reported on WSJ.}
		\label{tb:wsj_wer2}
		\begin{tabular}{l||c|c}
		     \hline
		     End-to-end ASR systems & dev93 & eval92 \\
		     \hline
		      seq2seq \cite{bahdanau2016end}  &    -      &  9.3  \\
		      CTC \cite{graves2014towards} &    -      &  8.2 \\
		      CTC \cite{miao2015eesen} &    -      &  7.3 \\
		      seq2seq \cite{chorowski2016towards}  &    9.7    &  6.7 \\
		      Multi-level LM (20K) \cite{hori2017multi}  & 9.6 & 5.6 \\
		      Look-ahead LM (65K) [this work] & {\bf 8.4} & {\bf 5.1} \\
		      \hline
		\end{tabular}
	\vskip -3mm
\end{table}

\subsection{Evaluation with LibriSpeech}
We conducted additional experiments using LibriSpeech to examine the performance of RNN-LMs including character LM, multi-level LM and look-ahead word LM for a larger corpus.
The data sets are summarized in Table \ref{table:librispeech}.
\begin{table}[t]
\caption{LibriSpeech data sets used for evaluation}
\label{table:librispeech}
\centering
\begin{tabular}{c|c|c}
    \hline
           & \# utterances & Length (h) \\
    \hline
    Train set  & 281,231 & 960 \\
    dev clean  & 2,703 & 5.3 \\
    dev other  & 2,864 & 5.1 \\
    test clean & 2,620 & 5.4 \\
    test other & 2,939 & 5.3 \\
    \hline
\end{tabular}
\vskip -3mm
\end{table}
All the experiments for LibriSpeech were performed using ESPnet, the End-to-End Speech Processing Toolkit \cite{watanabe2018espnet}, and the recipe for a baseline LibriSpeech setup with PyTorch backend \cite{ketkar2017introduction}.
According to the baseline recipe, we trained an 8-layer BLSTM encoder including 320 cells in each layer and direction, and the linear projection layer with 320 units followed by each BLSTM layer.
The second and third bottom LSTM layers of the encoder read every second state feature in the network below, reducing the utterance length by a factor of four, i.e., $T/4$. We also used location-based attention with a similar setting to the WSJ model.
The decoder was a one-layer unidirectional LSTM with 300 cells.
We also trained different language models as prepared for WSJ task, where we used only transcription of audio data including 9.4M words.
The both character and word RNN-LMs had 2 LSTM layers and 650 cells per layer. The beam width was set to 20 for decoding.

Table \ref{tb:librispeech_wer} shows word error rate (WER) with different language models. We obtained consistent error reduction with WSJ's results in Table \ref{tb:wsj_wer}, where
the both multi-level and look-ahead LMs provided significant error reduction when the vocabulary size was increased to 65K.
In this case, the look-ahead LM had competitive WERs to the multi-level LM.
However, the look-ahead LM still has the speed benefit similar to the results in Fig. \ref{fig:decoding-time} and the other benefit that we can completely exclude the training process of the character LM.
\begin{table}[tb]
	\centering
		\caption{Word error rate (WER) with different language models on LibriSpeech}
		\label{tb:librispeech_wer}
		\resizebox{0.48\textwidth}{!}{
		\begin{tabular}{l||c|c|c|c}
		     \hline
		     Language models (vocab. size) & dev & dev & test & test \\
		     		                     & clean & other & clean & other \\
		     \hline
		      No LM                    & 7.7 & 21.1 & 7.7 & 21.9 \\
		      Character LM             & 6.6 & 18.3 & 6.6 & 19.1 \\
		      Multi-level LM (20K)     & 5.7 & 16.0 & 5.9 & 16.8 \\
		      Look-ahead LM (20K) & 6.3 & 16.6 & 6.4 & 17.4 \\
		      Multi-level LM (40K) & {\bf 5.4}  & {\bf 15.6} & {\bf 5.5} & {\bf 16.5} \\
		      Look-ahead LM (40K) & 5.6 & 15.8 & 5.7 & 16.7 \\
		      Multi-level LM (65K) & {\bf 5.4}  & {\bf 15.6} & {\bf 5.5} & 16.6 \\
		      Look-ahead LM (65K) & {\bf 5.4} & {\bf 15.6} & {\bf 5.5} & {\bf 16.5} \\
		      \hline
		\end{tabular}}
	\vskip -3mm
\end{table}

\section{Conclusion}
In this paper, we proposed a word-based RNN language model (RNN-LM) including a look-ahead mechanism for end-to-end automatic speech recognition (ASR).
In our prior work, we combined character-based and word-based language models in hybrid CTC/attention-based encoder decoder architecture.
Although the LM with both the character and word levels achieves significant error reduction, two different LMs need to be trained and used for decoding, which increase the computational cost and memory usage.
The proposed method allows us to decode with only a word-based RNN-LM, which leads competitive accuracy and less computation in the beam search process compared to the multi-level LM approach.
Furthermore, it can completely exclude the training process for the character-level LM.
We have shown the efficacy of the proposed method on standard Wall Street Journal (WSJ) and LibriSpeech tasks in terms of computational cost and recognition accuracy. 
Finally, we demonstrated that the proposed method achieved 5.1 \%WER for WSJ Eval'92 test set when the vocabulary size was increased, which is the best WER reported for end-to-end ASR systems on this benchmark.
	
	
\balance

\bibliographystyle{IEEEbib}
\bibliography{strings,refs}

\end{document}